\def\year{2020}\relax
\documentclass[letterpaper]{article} 
\usepackage{aaai20}  
\usepackage{times}  
\usepackage{helvet} 
\usepackage{courier}  
\usepackage[hyphens]{url}  
\usepackage{graphicx} 
\usepackage{booktabs}

\usepackage{verbatim}
\usepackage{xspace} 
\usepackage{graphicx}
\usepackage{adjustbox}
\usepackage{array}
\usepackage{multirow}
\usepackage{caption}
\usepackage{subcaption}

\usepackage[ruled, linesnumbered]{algorithm2e}
\usepackage{graphicx}  
\usepackage{color}

\newcommand{\citet}[1]{\citeauthor{#1}~\shortcite{#1}}

\newcommand{\fair}{$^1$}
\newcommand{\ucl}{$^3$}
\newcommand{\loria}{$^2$}
\newcommand{\uclfair}{$^{1,3}$}
\newcommand{\loriafair}{$^{1,2}$}

\title{Generating Interactive Worlds with Text}
\author{{Angela Fan$^*$\loriafair{}, Jack Urbanek\thanks{~Joint first authors.}\fair{}, Pratik Ringshia\fair{}, Emily Dinan\fair{},} \\ {\Large\bf Emma Qian\fair{}, Siddharth Karamcheti\fair{}, Shrimai Prabhumoye\fair{}, Douwe Kiela\fair{},} \\ {\Large\bf Tim Rockt\"aschel\uclfair{}, Arthur Szlam\fair{}, Jason Weston\fair{}} \\  \fair{}Facebook AI Research \\
\loria{}LORIA, Nancy\\
\ucl{}University College London\\ \texttt{light-dms@fb.com}}

\begin{document}

\maketitle

\begin{abstract}

Procedurally generating cohesive and interesting game environments is challenging and time-consuming. In order for the relationships between the game elements to be natural, common-sense has to be encoded into arrangement of the elements. 
In this work, we investigate a machine learning approach for world creation using content from the multi-player text adventure game environment LIGHT \cite{urbanek2019learning}. We introduce neural network based models to compositionally arrange locations, characters, and objects into a coherent whole.  
In addition to creating worlds based on existing elements, our models can generate new game content. Humans can also leverage our models to interactively aid in worldbuilding. We show that the game environments created with our approach are cohesive, diverse, and preferred by human evaluators compared to other machine learning based world construction algorithms.
\end{abstract}

\section{Introduction}

A large component of fantasy and science fiction literature is \textit{worldbuilding}: putting together an elaborate context, with interesting (but believable) details, that can serve as a backdrop to a story (or for many stories). Successful worldbuilding requires common-sense knowledge about the real world and an understanding of the expectations of the audience.   

\begin{figure}[t!]
    \centering
    \includegraphics[width=\linewidth]{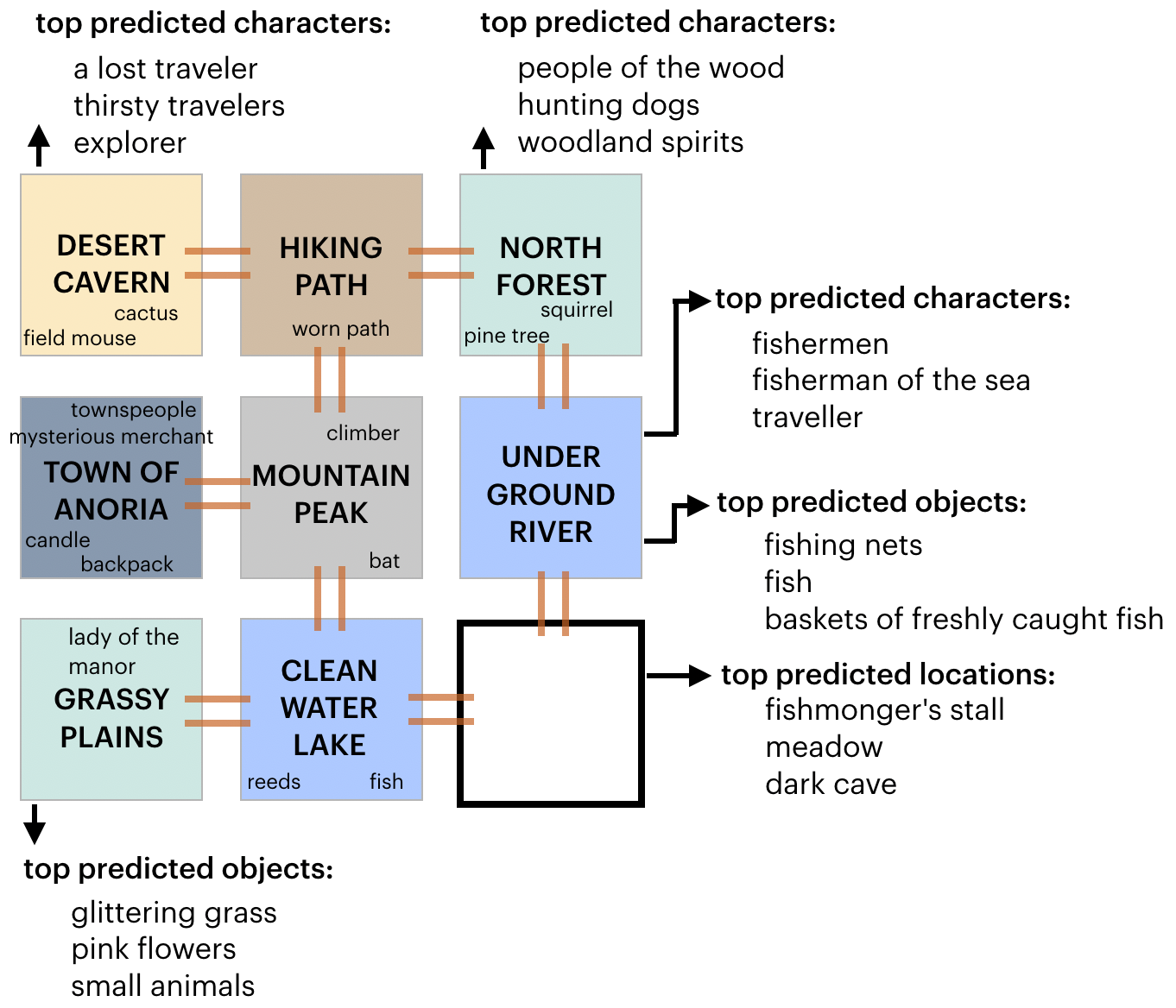}
    \caption{\textbf{Sample Constructed Game World}. Models arrange locations, then populate them with characters and objects. Top predictions are shown. Table~\ref{fig:light_example} shows the descriptions associated with the location \textit{Town of Anoria}, placed in middle left of this generated world.}
    \label{fig:light_example_predictions}
\end{figure}

In this work, we present a machine learning (ML) approach to creating a cohesive and interesting world built from elements of the text-based fantasy game environment LIGHT \cite{urbanek2019learning}.   These crowd-sourced elements, including descriptions of locations, characters, 
and objects, provide a rich source of supervision for learning common-sense relationships. Previous work in LIGHT focused on static, single-location settings using the crowd-sourced data. Instead, we focus on creating full environments for players to explore. We show how ML algorithms can learn to assemble these different elements, arranging locations and populating them with characters and objects. We use models to learn how to answer questions such as: \textit{Where is the ornate trunk likely to be? What is likely to be inside it? Where is the knight likely to be?} These considerations are necessary for building a cohesive game environment. 

We demonstrate that our proposed models can construct rich game environments that are diverse and preferred by human evaluators. We also develop models to generate descriptions of new locations, characters, and objects. Finally, we demonstrate that these machine learning tools can aid humans interactively in designing new game environments.

\begin{table}[t!]
  \begin{center}
    {\small
        \begin{subtable}{\linewidth}
        \centering
        \captionsetup{width=\linewidth}
          \begin{tabular}{l|l}
            \toprule
            \textbf{Name:} & Town of Anoria \\
            \midrule
            \textbf{Description:} & Town of Anoria has lots of cobble stone  \\ 
            & streets and wood houses of one story. [...]  \\
            & The town of Anoria is inland and takes a \\ 
            & long time to reach the sea, [...]\\
            \midrule
            \textbf{Neighbors:} & Mountain's Peak\\
            \midrule
            \textbf{Characters:} & townspeople, mysterious merchant \\
            \midrule
            \textbf{Objects:} & candle, backpack\\
            \bottomrule
          \end{tabular}
          \subcaption{\textbf{Example location}: Our model placed the Town of Anoria with an exit to the Mountain Peak, and placed characters and objects inside this location.\\}
          \label{table:world-room}
          \end{subtable}
          \newline
          \newline
          \begin{subtable}{\linewidth}
          \centering
          \begin{tabular}{l|l}
            \toprule
            \textbf{Character:} & Mysterious Merchant \\
            \midrule
            \textbf{Persona:} & I am the mysterious merchant of the village. \\
            & I sell rarities from around the world  \\
            & that can not be purchased anywhere else. [...] \\ 
            \midrule
            \textbf{Description:} & The merchant in town came and went \\ 
            & without a crack of the grass beneath his feet. \\
            & No one knew when he was gone, nor when \\ 
            & he returned home [...] \\
            \midrule
            \textbf{Carrying:} & pouch, cane \\
            \midrule
            \textbf{Wearing:} & hat, coat\\
            \midrule 
            \textbf{Wielding:} & dagger \\ 
            \bottomrule
          \end{tabular}
          \subcaption{\textbf{Example character}: Our model placed the Mysterious Merchant in the Town of Anoria, along with other townspeople.\\}
          \label{table:world-characters}
          \end{subtable}  
          \begin{subtable}{\linewidth}
          \centering
          \begin{tabular}{l|l|l}
            \toprule
            \textbf{Object} & \textbf{Description} & \textbf{Affordances}\\
            \midrule
            pouch & The pouch is made of fine silk & container \\ 
            & cloth, colored bright red. It has a   & gettable \\ 
            & leather string keeping it sealed. &  \\
            \midrule
            cane & The cane is made of a very  & gettable \\
            &  uncommon, ornate wood. & \\ 
            \midrule
            dagger & The dagger is curved & gettable \\ 
            & with a golden hilt. & weapon \\
            \bottomrule
          \end{tabular}
          \subcaption{\textbf{Example objects}: Pouch, Cane, and Dagger, all carried by the Mysterious Merchant.\\}
          \label{table:world-objects}
          \end{subtable} 
          
          \begin{subtable}{\linewidth}
          \centering
          \begin{tabular}{l|l}
            \toprule
            \textbf{Object} & \textbf{Inside the Object}\\
            \midrule
            pouch & coins, eyeglasses \\
            backpack & wallet, bedroll, tools \\
            \bottomrule
          \end{tabular}
          \subcaption{\textbf{Example objects within container objects}: Our model placed additional objects inside the Pouch and the Backpack.}
          \label{table:world-objectsIn}
          \end{subtable} 
          \caption{Game Elements include locations, characters, objects, and objects within containers. Elements have descriptions and annotations such as what a location contains.
     \label{fig:light_example}}
    }
  \end{center}
\end{table}

\section{Constructing Game Environments}
\label{sec:const}

In this section, we detail methods for learning to build game elements from compositions of sub-elements; and worlds from these elements.

\subsection{Background on LIGHT}
LIGHT is a multi-player text-based fantasy-themed virtual world. It consists of a set of crowd-sourced \textit{game locations, characters, and objects}, and a game engine that controls the interactions between these.  Characters can speak to each other via text, send emotes like \emph{grin} or \emph{ponder}, and take actions to move to different locations and interact with objects.  Some example actions include
{\em go north},  
{\em get shovel}, or {\em unlock door}.
The game engine represents the game state as a graph, and the actions by characters amount to operations on the graph. 
The locations, characters, and objects were crowd-sourced using Amazon Mechanical Turk. Crowd-workers were asked to provide names and descriptions for each of these aspects through natural language, for a total of 663 locations, 1755 characters, and 3462 objects. See Table~\ref{fig:light_example} for examples and Figure~\ref{fig:light_example_predictions} to see how our work combines the elements into a playable game environment.


\subsection{Building a Game World}
\citeauthor{urbanek2019learning} focused on modeling character dialogue and action in pre-built locations. ML models were trained to play the game by mimicking the actions and dialogues of human players in fixed settings built by crowd-workers. In contrast, in this work, we study models for assembling the game itself rather than agents that play it.  Since these elements were separately crowd-sourced, we can compose them to create a large number of different game environments. 

We describe our approach from the top down. 
First, in Section \ref{sec:locations} we discuss connecting pre-built locations together to form a world. In \ref{sec:building_rooms}, we give our methods for filling a location with characters and objects. We describe the additional data collected to model objects \textit{contained} within other objects. Next, in \ref{sec:building_elements}, we discuss how to generate new game elements.  In Section \ref{sec:interactive} we describe how these methods can be modified and utilized for interactive world-building. Finally, in Section~\ref{sec:together}, we describe how to bring these models together to create a new game world.

\subsection{Building Maps by Arranging Locations} 
\label{sec:locations}

We describe our method to train machine learning models to arrange locations in LIGHT.

\paragraph{Locations in LIGHT} Each \textit{location} represents a place  with a \textit{name} and \textit{description}. The description provides background information about the location and what a player might see as they enter it. Crowd-workers provided examples of neighboring locations, as well as what characters and objects would be present within the location.

\paragraph{Using Machine Learning to Place Locations} Game locations must be spatially arranged so as to create a logical and cohesive environment for players to explore. For example, the \textit{Wizard's Reagent Room} being located near the \textit{Wizard's Tower} would make a more intuitive game experience compared to locations being randomly placed.  

To train models for this task, 
we use the example neighbors for each location provided by crowd-workers, obtaining triplets of (location name, location description, location neighbors). We partitioned this into a training, validation, and test set such that the locations are distinct in each set (see Table~\ref{table:task_stats}). As each location can have multiple neighbors, the individual datapoints available for the prediction task is larger than the number of total locations collected. 

We consider a variety of different \textit{ranking models} for this task, in two settings. In the first, models have access to the location name only, and in the second, they additionally have access to the location description information. These models compare the human annotation of  neighboring locations with a variety of negative candidates. These negative candidates can be thought of as distractor locations from the dataset that the model must distinguish from the human annotated location, similarly to how negative training data is sampled in the knowledge base population literature \cite{bordes2013translating}. Models are trained to maximize the score of the human response and minimize the scores of the negative candidates. When constructing a new world at test time, the placed location is the highest scoring candidate from the model prediction. We use two  machine learning approaches:
\begin{itemize}
    \item \textit{Starspace:} The Starspace \cite{wu2018starspace} model  learns a bag-of-words embedding for the location information (e.g. name and description). The model encodes the location information as well as the negative candidates, and trains to maximize the inner product of the true human annotation. We initialized the Starspace model using \textit{fasttext}, a method for learning vector representations of individual words. This initialization allows the model to begin training with a better understanding of the text.
    \item \textit{BERT-based Models:} Recent work \cite{devlin2018bert} in natural language processing has shown strong performance of the BERT model, which learns to encode text in a left-to-right and right-to-left fashion by training on large quantities of text data available online. We use the BERT-based models proposed in \cite{urbanek2019learning,humeau2019real} to encode the location information and the negative candidates. We explore two variants: 
    
    (1) \textit{Bi-Encoder}, which encodes the candidates and input context separately. This model scores the candidates by calculating the dot product between these embeddings. 
    
    (2)  \textit{Cross-Encoder}, which concatenates the context with each candidate before encoding, allowing this model to build a context-dependent representation of each candidate. This model scores candidates by projecting the vector representation of text to a scalar.
\end{itemize}

As we have a limited quantity of data for the task, we found that using input dropout to prevent overfitting was crucial for good performance for both of these models.  

We compare these models that learn from the training data with three baselines:
\begin{itemize} 
    \item \textit{Random:} We report a random baseline that selects a random candidate from the provided negative candidates.
    \item \textit{Data Proportional:} Instead of selecting candidates fully at random from the provided negative candidates, we select proportional to the number of times that candidate appears in the training set. This leverages the data annotation information and reflects that some candidates are more likely to be used than others. 
    \item \textit{Information Retrieval:} This model selects the candidate with the largest word overlap using TF-IDF weighting.
\end{itemize}

To create a map for a new game, models are used to predict the neighboring locations of each existing location. For each new location added, the model will fill in the surroundings. A location can connect to up to four neighboring locations, though not all connections need to be filled. To make the game environment more interesting and diverse, locations cannot appear multiple times in one map (e.g. \textit{Berka's Forest Inn} is only located in one place). 

\paragraph{Adding Filler Locations} A challenge with using crowd-sourced data for all of the locations is that crowd-workers often write exciting and complex locations. However, when players explore the game environment, this tendency leads to each location being complex and overwhelming. To remedy this, we create a set of 25 filler locations such as \textit{abandoned shack}, \textit{empty closet}, and \textit{storage room} that provide additional content between the exciting locations that crowd-workers described. Filler locations can appear multiple times (e.g. there can be multiple \textit{empty closets}).

\subsection{Adding Characters and Objects to Locations}
\label{sec:building_rooms}

We describe how to apply our methods to add characters and objects to predicted locations.

\paragraph{Characters in LIGHT} Each character is described by a \textit{name}, \textit{persona}, and a \textit{description}. The persona provides information about the character, such as their background and motivation, while the description describes the character's appearance.  LIGHT also has annotations of objects characters would carry, such as a \textit{Wizard} holding a \textit{staff}.  

\paragraph{Objects in LIGHT} Each object represents an item that characters can interact with, such as \textit{get shovel}. Objects have a \textit{name}, a \textit{description}, and a set of \textit{affordances}. The description lists what the object looks like and what it might be able to do. The affordances represent object properties, such as gettable and drinkable. These are used by the game engine to determine the set of possible interactions of the object. For example, objects with the drinkable affordance can be interacted with using the action \textit{drink}. Objects can be inside other objects, to represent for example \textit{coins inside a wallet}. We crowd-sourced additional annotations of object size and examples of other objects that could be inside. 

\paragraph{Using Machine Learning to Place Characters and Objects}
\label{sec:ml_for_obj_car}

Using the characters and objects associated to locations from LIGHT as ground-truth, we create training, validation, and testing data (see Table~\ref{table:task_stats}) 
to fit models to place characters and objects in locations, as well as object within objects. 
To collect objects within objects data, crowd-workers were given an object with the container affordance and asked to name multiple objects that could be inside. 

We place characters and objects using the models described in Section~\ref{sec:locations}. Here, instead of predicting neighboring locations, models are given locations and trained to predict characters and objects, or given objects and trained to predict which objects could be inside. For example, the \textit{character prediction} task would receive as input the location \textit{Wizard's Reagent Room} and  predict \textit{Wizard}. As the amount of data for each task is low, we employ multi-task learning and train all of the tasks (locations, characters, objects, and  containers) together to increase the quantity of training data.

\begin{table}[t]
\begin{center}
\small
\begin{tabular}{l|ccc}
 \toprule
Split                & Train & Valid & Test \\
\midrule                
Locations             & 914   & 109  & 110    \\
Characters           & 529  &  305  &  305 \\
Objects              & 359  &  318  &  256  \\
Object Containers   &  359  & 318  &   256   \\
\bottomrule
\end{tabular}
\caption{\textbf{Dataset Statistics for World Generation:} arranging locations next to each other, placing characters and objects within locations, and placing objects within objects.
\label{table:task_stats}
}
\end{center}
\end{table}

\subsection{Generating New Game Elements}
\label{sec:building_elements}

Adding new elements to the existing LIGHT game is complex: descriptions, object affordances, character personas, and other details would need to be written. Instead, we propose using generative machine learning models to create  additional content based on the name of the new item (either a location, a character, or an object). We use the same training, validation, and testing splits used in the world construction task (see Table~\ref{table:task_stats}). These generated items can be added to the game environment, so newly generated game worlds can incorporate them along with existing crowd-sourced elements. 

We use the Transformer \cite{vaswani2018attention} neural network architecture to create a Sequence-to-Sequence model to make the following predictions: 
\begin{itemize}
    \item Given location name, predict background and description
    \item Given character name, predict persona and description
    \item Given object name, predict description and affordances 
\end{itemize}

We compare the Transformer in two settings: with and without \textit{pretraining}. As the dataset for generating new game elements is small, the generative model can be trained on a larger corpus and finetuned on this task. We use a large dataset of 2 billion Reddit comments for pretraining. Reddit comments are chosen because they are close to natural human conversation and exhibit elements of creativity and story-telling that may help generate interesting descriptions. 

To be able to handle new vocabulary and ease learning, we use byte-pair encoding \cite{sennrich2016neural} to model subwords. Similar to Section~\ref{sec:ml_for_obj_car}, we multi-task prediction location, character, and object description, location background, and character persona with one model. We use top-$k$ sampling \cite{fan2018hierarchical} to reduce repetition during generation. Object affordances are predicted with a separate model, as multi-label classification between seven possibilities is distinct from the other tasks. 

\subsection{Aiding Human Game Design}
\label{sec:interactive}
Machine learning models can be applied to automatically generate game environments for players, but they can also be used to aid humans in game design. Many existing game engines assist in fast and intuitive creation of different worlds already, for example providing level design tips or improving pathfinding \cite{graham2003pathfinding}. 
Our methods can be used to automatically suggest neighboring locations or which characters and objects to place in the existing locations, speeding up world design. 

\subsection{Proposed Algorithm for World Generation}
\label{sec:together}

How do we use our proposed models collectively to make a new game world? First, an empty map grid is initialized to represent the number of possible locations. A percentage of grid positions are marked inaccessible to make exploration more interesting. The central location is populated randomly. We use the best performing model to iteratively fill in neighboring locations until the entire grid is populated. Then, for each placed location, the model is used to predict which characters and objects should populate that location. Finally, the model is used to predict if objects should be placed inside existing objects. Figure~\ref{fig:light_example_predictions} displays an example generated world, with model predictions shown for missing elements. See Appendix for further details.

In an interactive setting where players are able to design their own worlds, we use models to provide suggestions for which elements to place. If players enter names of game elements not present in the dataset, our generative models are used to write descriptions, personas, and affordances.

\section{Related Work} 
\label{sec:related_work}

\begin{figure}[t!]
    \centering
    \includegraphics[width=0.8\linewidth]{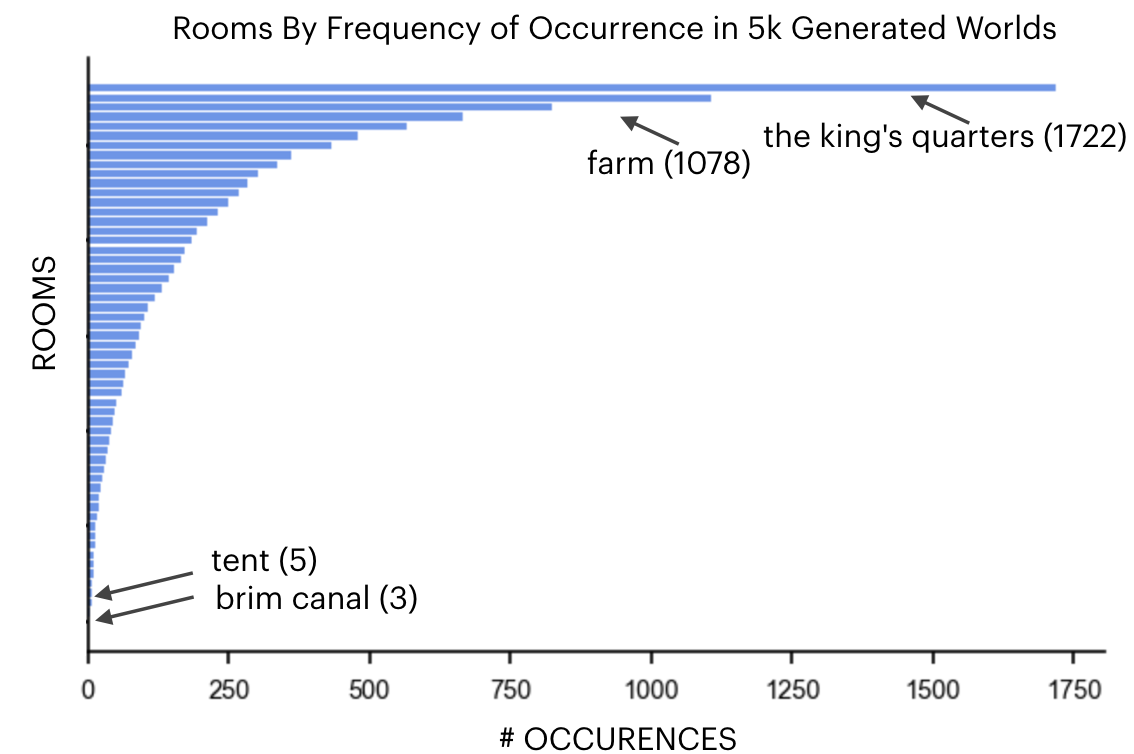}
    \caption{\textbf{Frequency of Location Placement} in 5000 generated game environments using our models.}
    \label{fig:room_examples}
\end{figure}

\begin{figure*}[t!]
    \centering
    \includegraphics[width=0.7\linewidth]{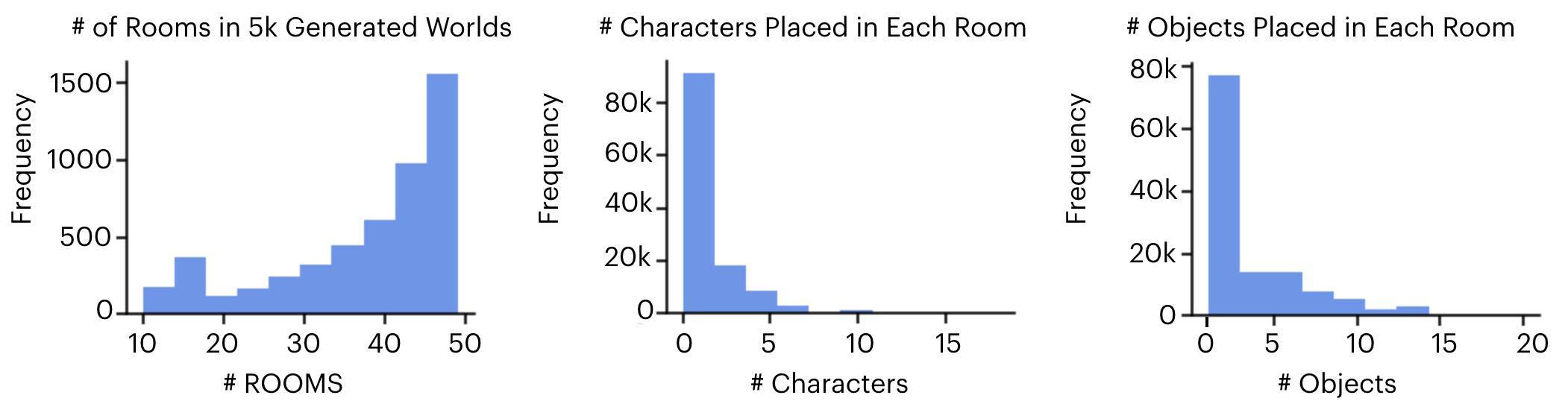}
    \caption{\textbf{Distribution of Locations, Characters, and Objects in 5,000 generated maps}. Our method generates fairly large maps (the maximum size is set to 50) and places 1-3 characters and objects in each location.}
    \label{fig:room_freq}
\end{figure*}

\begin{figure*}[t!]
    \centering
    \includegraphics[width=0.7\linewidth]{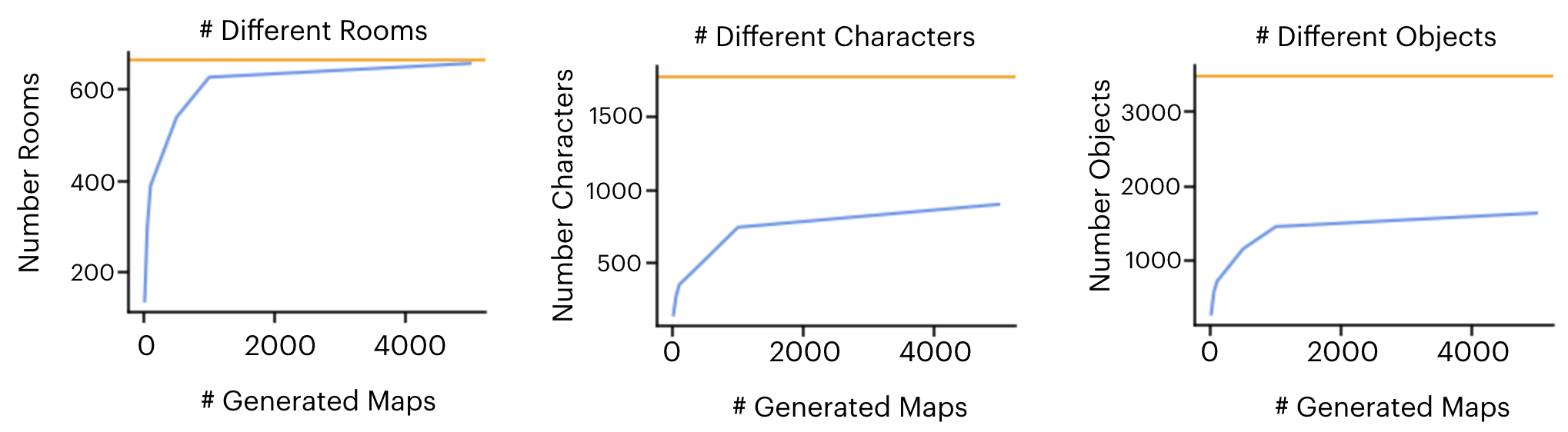}
    \caption{\textbf{Number of Different Locations, Characters, and Objects as a Function of Generated Maps}. As additional maps are generated, a greater diversity of game elements appears. The orange line denotes the total number of elements in the dataset.}
    \label{fig:room_sets}
\end{figure*}

\paragraph{Procedural Content Generation in Games}

Using algorithms to aid game generation is a growing field as the popularity of gaming rises. Recent work has made  progress on level design in various game settings \cite{guzdial2018automated,khalifa2016general,summerville2016learning,van2013designing,vara2014creating}, including rhythm games \cite{lin2018generationmania}, physics games \cite{stephenson2016procedural}, dungeon exploration  \cite{shaker2016constructive}, and social games \cite{risi2012combining}. 

Much prior work has focused on the task of level generation, but there are other facets of games that could be generated. For example, weapons, various items, and characters are present in game levels \cite{liapis2014computational,liapis2018orchestrating}. We focus on how the various facets could fit together within a text-based game, and how we can use them to generate an entire game environment.

\paragraph{Text-Based Games} Many settings for content generation in text-based games have been explored. For example, \citet{barros2016murder} use text from Wikipedia to link various entities for the generation of murder mystery games. \citet{ammanabrolu2018playing} represent a text-based adventure game as a graph and learn how to adventure within this world. Work has been done to generate Sporcle-like textual quizzes.\footnote{https://github.com/vjuylov/txt2lvl} Our work is similar to \citet{cote2018textworld}, which introduces a framework for  procedurally generating text-based environments that can be used as environments for learning agents. 
They find room configurations based on the number of desired rooms and overall map size, and these rooms are then populated with objects via human-designed algorithms. In contrast, our work focuses on using machine learning algorithms to create cohesive environments by either composing crowd-sourced components or through generative models.

\paragraph{Generation using Machine Learning}

Generative modeling is an important topic in machine learning, outside of games or other creative endeavors.  Recent works have demonstrated impressive models of images \cite{karras2017progressive} and text \cite{radford2019language}. Recently, statistical ML has also been proposed for creative endeavors. For example \citeauthor{gatys2016image} show how users can manipulate the style of images using convolutional neural networks and \cite{zhu2017your,sbai2018design} describe ML-aided fashion design. There has been work in ML for music generation, see e.g. Magenta\footnote{\url{https://magenta.tensorflow.org/}} or \citet{briot2017deep} for a survey.  Most related to our  world construction are methods for generating stories, poetry, and scripts \cite{fan2018hierarchical,ghazvininejad2016generating,janghorbani2019domain,marti2018cardinal}. 

Work in content generation with machine learning has incorporated human guidance. For example, several works incorporate human control such as length and style to improve summarization, dialogue, and text simplification \cite{fan2018controllable,see2019makes,martin2019controllable}. \citet{wang2018high} generates portions of images after human editing.  

\section{Evaluation and Results}
\label{sec:eval}

We discuss several evaluations of both elements and worlds, compare methods, and discuss their successes and failures. 

\subsection{Diversity of Generated Worlds}
 \label{sec:world_diversity}
 
Our proposed method can be used to automatically create a variety of diverse game worlds. We generate 5,000 worlds with a maximum size of 50 arranged locations and analyze these generations to understand the properties of created game environments. 

\paragraph{Locations} The generated maps are very diverse. Figure~\ref{fig:room_sets} shows the number of map generations required to generate the full number of locations in the dataset. With 500 generations, a large majority of different locations have been used. Over 95\% of locations in the dataset are used after 5000 generations. The most commonly placed location is \textit{the king's quarters}, in  34\% of the generated worlds (see Figure~\ref{fig:room_examples}). Some locations are used very sparingly, such as the \textit{brim canal} (0.06\% of the worlds). Allowing our modeling approach to decide the map size, 80\% of the generated worlds have more than 30 locations (see Figure~\ref{fig:room_freq}) and about 40\% of the worlds have the maximum number locations. Example generated maps are shown in the Appendix.

\paragraph{Characters} Around 65\% of characters in the dataset are generated after 5000 maps (Figure~\ref{fig:room_sets}). The lower coverage is most likely because there are very specific characters created by crowd-workers that are not scored highly by models, and thus not often placed. To provide a concrete example, a specific qualified character might be in the dataset, such as \textit{an old, wizened priestess}, but if that character is only mentioned once in the training set, a model might score a more generic character higher, such as \textit{priestess}. The maximum number of characters placed in one room is around 15, but most locations have 0-3 characters present (see Figure~\ref{fig:room_freq}).\footnote{ML approaches are known to reflect data biases  \cite{zhao2019gender,brunet2018understanding}. We found that there are a greater number of male characters in LIGHT, and this is reflected in the generated environments. We plan to investigate this in a follow-up work.} 

\paragraph{Objects} Similar to characters, around 60\% of objects in the dataset are generated after 5000 maps, shown in Figure~\ref{fig:room_sets}. Some locations contain a large number of objects, such as the \textit{Treasure Chamber}, but most locations contain about 1-3 objects that players can interact with (Figure~\ref{fig:room_freq}).

\begin{table*}[t!]
  \centering \small
  \begin{tabular}{l l|  c c c c }\toprule
    \bf{Feature} & \bf{Model} & \textbf{Locations} & \bf{Characters} & \bf{Objects} & \bf{Containers}\\\midrule
    &   Random & 8.2 & 5.9 & 5.9 & 5.7 \\ 
    &   Data Proportional &  0.0  &  9.8   &    20.1 &  5.9   \\ \midrule
    Name Only &   Information Retrieval & 18.2 & 7.5 & 8.2 & 9.6 \\ 
    &   Fasttext & 9.1 & 12.8 & 15.6 & \textbf{27.4} \\ 
    &   Starspace & 44.5 & 17.7 & 13.3 & 20.1 \\ 
       \midrule 
    Name and Description &   Information Retrieval & 30.0 & 19.0 & 21.9 & --- \\
    &   Fasttext & 28.2 & 17.0 & 16.8 & --- \\ 
    &   Starspace & \textbf{45.5} & 35.7 & \textbf{47.3} & --- \\ 
    &   BERT Bi-Encoder & 30.2 & 30.2 & 34.0 & --- \\ 
    &   BERT Cross-Encoder & 28.2 & \textbf{36.1} & 35.5  & --- \\ 
        \bottomrule
\end{tabular}
   \caption{\textbf{Comparison of Various Approaches to Worldbuilding.} We report Hits at 1 on the test set for arranging locations and populating with objects, characters, and placing objects within container objects. Starspace models perform well on all tasks.}
 \label{tbl:room_fill_experiments}
\end{table*}

\subsection{Quality of Generated Worlds}
 \label{sec:world_quality}
\paragraph{Automatic Evaluation} We first use \textit{automatic evaluation} to compare the quality of different machine learning approaches to the location, character, object, and container prediction tasks. We measure \textit{Hits at 1}, or the percentage of time the correct candidate is ranked first amongst the negative candidates. If the model always predicted  what the crowd-workers annotated, then this metric would have the value 100.
Containers are evaluated in the \textit{Name only} variant --- as crowd-workers were able to write any object, not all of their written choices have descriptions.

Results are shown in Table~\ref{tbl:room_fill_experiments}. Leveraging the data distribution to weight random sampling provides a strong baseline for characters and objects, as a few are quite common. Providing the description text is helpful for improving prediction quality compared to having access only to the name feature. Amongst the various approaches, Starspace models show strong performance, particularly on the location prediction task. The Bi and Cross Encoder models are very large neural networks and may be overfitting on the much smaller LIGHT world creation training data. Further, they are pretrained on non-domain specific data, which may negatively impact performance.

\paragraph{Human Assessments} We conduct \textit{human evaluation} to compare the various approaches to world generation. We compare the performance of two models pairwise by starting in the same location and using the models to iteratively predict subsequent locations. Locations are then populated with characters and objects. After each location, human evaluators are asked which model was able to place more logical and interesting characters and objects. After five steps through predicted locations, human evaluators are asked which model path they prefer as more natural, cohesive, and interesting. We compare four different approaches:

\begin{itemize}
    \item \textit{Random:} Locations, characters, and objects were randomly selected from the set of all possible datapoints.
    \item \textit{Starspace:} The model described in Section 2 was used to predict which locations should be linked in the path, and the characters and objects present in each location.
    \item \textit{Data Created Paths:} This method uses the existing dataset of annotated locations and their neighbors to construct a path. The characters and objects present in each location are from the original crowd-sourcing tasks. In contrast to Starspace, the number of possible paths that could be created with existing data is limited. For example, if a room has only one annotated neighbor, it would always be arranged in the same manner.
    \item \textit{Human Annotated Paths:} Human evaluators constructed paths by manually linking locations. Here a single evaluator created an entire path --- in contrast, in \textit{Data Created Paths}, annotators during initial data collection only provided a one-step neighbor, rather than one person creating the entire path. The characters and objects are from the original crowd-sourcing. While human-created paths could be high quality, such a method does not scale to large worlds as it is costly and time consuming.
\end{itemize}

\begin{figure}[t!]
    \centering
    \includegraphics[width=0.7\linewidth]{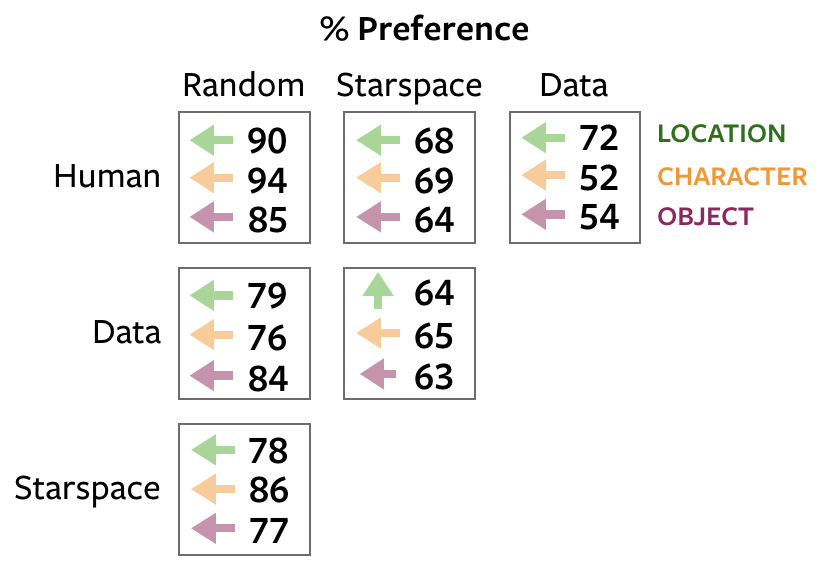}
    \caption{\textbf{Human Evaluation of World Construction}. The number indicates percentage preference, with the arrow pointing to the winner. The first row in each box is location preference, second character, and third object.}
    \label{fig:human_preference}
\end{figure}

As shown in Figure~\ref{fig:human_preference}, human evaluators prefer Human Annotated Paths the most, but Starspace prediction models perform strongly as well. Starspace is strongly preferred over Random and the predicted location paths are preferred over Data Created Paths over 60\% of the time. 

\begin{figure}[t!]
    \centering
    \includegraphics[width=0.7\linewidth]{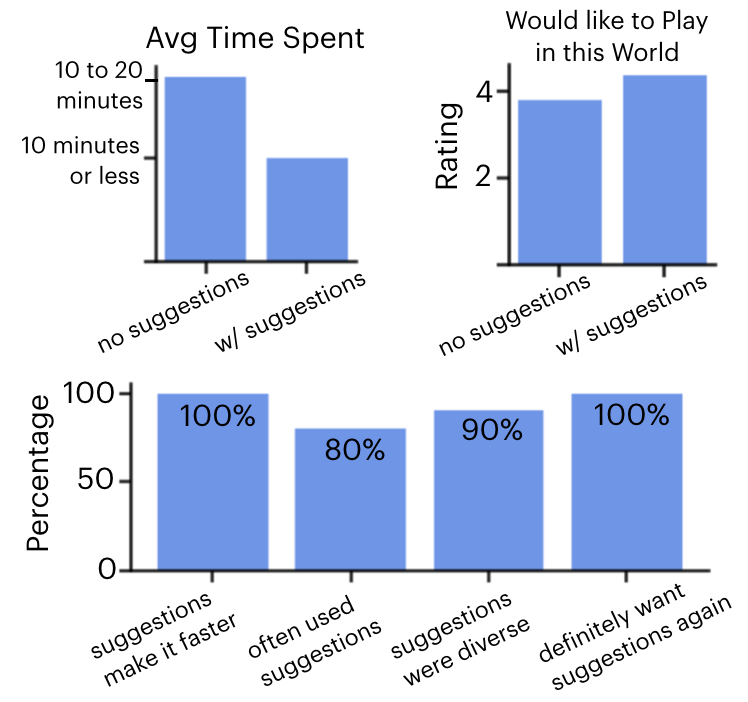}
    \caption{\textbf{Machine Learning Models Aiding Human Game Design.} Having access to model suggestions for which elements to place next allows evaluators to create fun and interesting game environments more quickly.}
    \label{fig:map_generation}
\end{figure}

\begin{table}[t!]
  \centering \small
  \begin{tabular}{ l | c c c | c}\toprule
    \bf{Model} & \bf{Location} & \bf{Character} & \bf{Object} & \bf{Affordance}\\\midrule
    - Pretrain & 23.3 & 28.4 & 32.5 & 57.0 \\ 
    + Pretrain & 26.3 & 29.3 & 38.8 & 62.6 \\ 
    \bottomrule
\end{tabular}
   \caption{\textbf{Comparison of Game Element Generation}. We report F1 on the test set.}
 \label{tbl:generation_experiments}
\end{table}

\begin{table}[t!]
  \centering \small 
  \begin{tabular}{l|l}
    \toprule 
    \textbf{Given Location:} & The forest \\
    \midrule
    \textbf{Generated} & This forest has been a part of the \\ 
    \textbf{Background:} & kingdom for centuries. It is a great \\
    & resource to have when needed. Most \\
    & people consider it a necessity to have a \\
    & peaceful and peaceful environment. \\
    \midrule
    \textbf{Generated} & The forest is filled with wild animals \\
    \textbf{Description:} & that live in the forest, but are unable \\ 
    & to climb up trees. It is clear the trees are \\
    & not in use. \\
    \toprule
    \textbf{Given Character:} & A Princess \\
    \midrule
    \textbf{Generated} & I love to help people and I love my \\ 
    \textbf{Persona:} & family. I spend my days fighting, \\
    & protecting my people and my children. \\
    & I love to help the royal family!\\
    \midrule
    \textbf{Generated} & She is a fierce looking princess that has\\
    \textbf{Description:} & been trained to protect any people. \\  
    \toprule 
    \textbf{Given Object:} & wooden sword \\
    \midrule
    \textbf{Generated} & This wooden sword is worn down and \\
    \textbf{Description:} & damaged, to say the least, it is a sword \\
    & that would be useful for a knight's duty. \\
    \midrule
    \textbf{Generated} & gettable  \\
    \textbf{Affordances:} & weapon \\ 
    \bottomrule
  \end{tabular}
  \caption{\textbf{Generations of New Game Elements}, given the desired name. Examples are chosen from the test set.}
  \label{tbl:generated_examples}
\end{table}    

\subsection{Generation of New Game Elements}
 \label{sec:eval_game_elements}
To evaluate the quality of automatically generating new game elements using our proposed models, we compare F1, a metric of word overlap. For this metric, the text is lowercased and the overlap between tokens is computed. Pretraining increases the performance on all generation tasks, as shown in Table~\ref{tbl:generation_experiments}. For character descriptions and personas, the effect of pretraining is minimal. We hypothesize this is due to the slightly more templated nature of written personas, as many begin with \textit{I am a}. Example generations are shown in Table~\ref{tbl:generated_examples}.
Our generative models are able to write interesting, new, and generally coherent descriptions for a variety of different game elements (see Appendix for additional examples).  We analyze the n-gram overlap of our generated game elements with the training set to understand how much of the written text is novel. We find that 34\% of generated 3-grams are present in the training set (largely common phrases), but only 2.5\% of generated 5-grams are present in the training set. As we are generating text with top-k sampling, the models do not tend to copy long sequences.  
          
\subsection{ML-aided interactive world creation}
 \label{sec:eval_interactive_creation}
To quantify if models can aid players in designing their own worlds, evaluators designed a nine-location game environment. Evaluators were explicitly told the goal was to make a text-based game interesting and fun. To add game elements, they have access to a search bar with autocomplete, so they can type what they wish to place and select from a list (see Appendix for an image of the user interface). Half of the evaluators have access to model predictions, which are surfaced as suggestions at the top of the search dropdown. However, they can choose to ignore the suggestions.

Evaluators created 10 game environments with access to model suggestions and 10 without. While they reported similar satisfaction with the diversity and quality of their generated worlds in both settings, the amount of time spent was different. Evaluators spent 10 minutes or less to create maps with suggestions and 10-20 minutes without suggestions (Figure~\ref{fig:map_generation}, top left). Those with suggestions said they would want to play an actual game in their created world more (Figure~\ref{fig:map_generation}, top right). Finally, evaluators had a positive reaction to model suggestions (Figure~\ref{fig:map_generation}, bottom): 100\% of evaluators agreed that suggestions made it faster to create a world and definitely would want to have them again, 80\% said they often chose from suggestions, and 90\% said the suggestions were diverse. Freeform feedback was positive, with comments such as \textit{suggestions foster creativity} and \textit{especially for characters, the suggestions showed what I wanted}. Additional results are shown in the Appendix.

\section{Conclusion} 

We proposed a method to procedurally generate game environments by using machine learning algorithms to arrange locations, place characters and objects within those locations and objects within containers, and write descriptions for new game elements. We explored different neural network based models for these tasks, and show with various automatic metrics and human studies that the maps generated by our approach are cohesive, interesting and diverse. Finally, we show that our machine learning approach can be used to aid humans in creating game worlds as well. Together, these steps show a path to creating cohesive game worlds from crowd-sourced content, both with model-assisted human creation tooling and fully automated generation. 

\bibliography{emnlp-ijcnlp-2019}
\bibliographystyle{aaai}

\clearpage
\newpage 
\section{Appendix}

\subsection{Pseudocode for Assembling a LIGHT World}

In Algorithm~\ref{alg:world_creation}, we denote in detail how to create a new, playable game environment for LIGHT using our proposed methods. 

\subsection{Model Details}

\paragraph{Starspace} Models were trained with embedding size 128 and embedding norm 10, initialized with fasttext embeddings. We trained with learning rate 0.01 and input dropout 0.5. We modeled a vocabulary of 10749 tokens. 

\paragraph{Transformer} Bi-Encoder and Cross-Encoder models leverage the BERT model \cite{devlin2018bert}. We finetuned them on the LIGHT tasks by warming up for 200 updates. We truncate the contexts and labels to 300 tokens. We train with input dropout 0.5.

\paragraph{Generative Transformer} contains 8 encoder layers and 8 decoder layers with 16 attention heads and 2048 FFN size. We model a BPE-based vocabulary of 54940 tokens. The text is truncated at 512 tokens. We optimize perplexity using Adam. 

\subsection{ML-Aided Game World Creation}

\begin{figure}[t]
    \centering
    \includegraphics[width=0.8\linewidth]{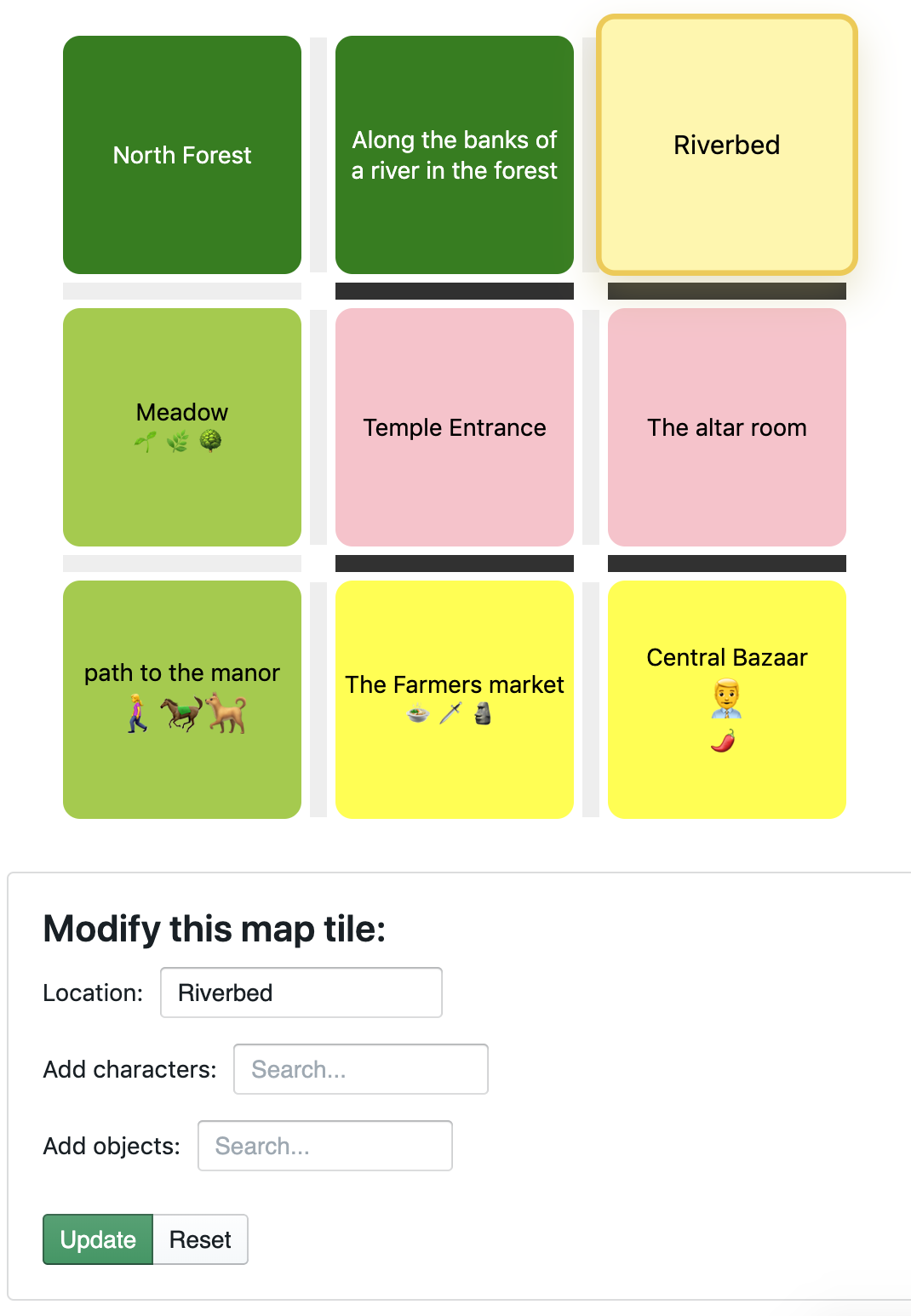}
    \caption{\textbf{User Interface for Map Creation}}
    \label{fig:world_creation_ui}
\end{figure}

The user interface that evaluators had access to is depicted in Figure~\ref{fig:world_creation_ui}. Evaluators were shown a grid of nine locations, with the center location populated randomly. This was done to give evaluators a starting point, and to encourage generation of diverse worlds. Evaluators can click on a location to highlight it (in Figure~\ref{fig:world_creation_ui}, the upper right location has been highlighted). Then, evaluators can use the different search bars to add locations, characters, and objects respectively. Once a map tile has been filled, the name of the location is labeled and the color changes according to the category of the location. For example, forests are colored green. We additionally mapped each character and object to an emoji, so the map tile would have a visual depiction to remind evaluators what they have already placed. For example, the \textit{Central Bazaar} location in the bottom right of Figure~\ref{fig:world_creation_ui} has a shopkeeper as a character and spices as an object.

\begin{figure*}[t]
    \centering
    \includegraphics[width=0.7\linewidth]{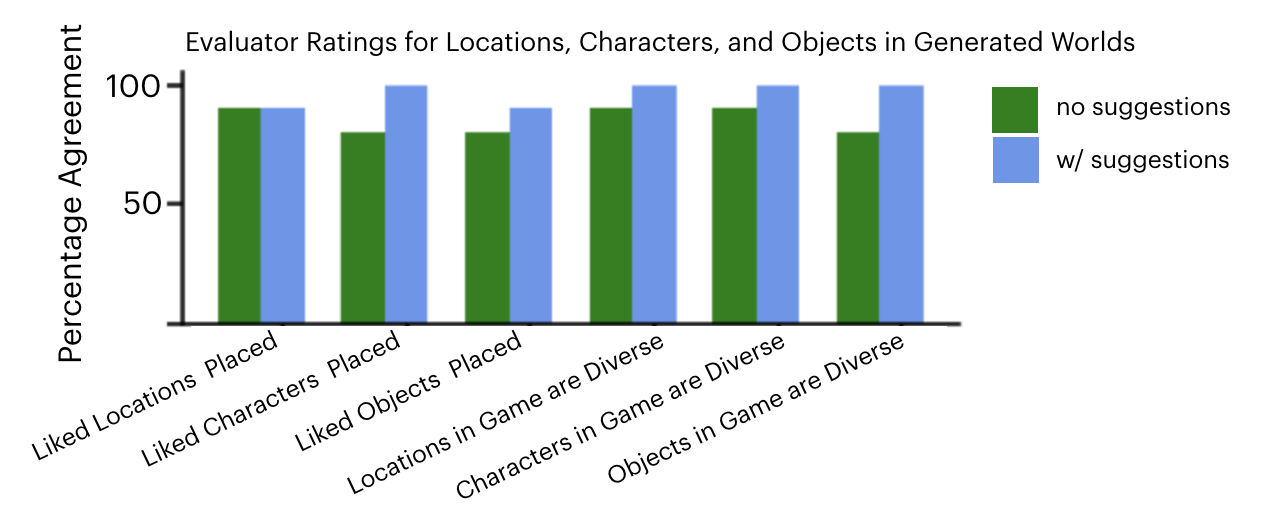}
    \caption{\textbf{Additional Survey Question Results.} Evaluators with access to model suggestions liked their placed characters more than evaluators without model suggestions, and rated the game locations, characters, and objects more diverse. Likely having the model suggestions allows evaluators to read a greater diversity of game elements.}
    \label{fig:survey_more}
\end{figure*}

We present results for various additional survey questions in Figure~\ref{fig:survey_more}. As shown, evaluators with access to model suggestions self-reported that they found the locations, characters, and objects more diverse- likely as the model surfaces suggestions they may not have thought of. Further, evaluators rated liking their placed characters substantially more with access to the model than without.

The list of survey questions we asked is the following:
\begin{itemize}
    \item How much time, in minutes, did you spend creating this world?
    \item On a scale of 1 to 5, with 5 being the best. How satisfied are you with the map you have built?
    \item On a scale of 1 to 5, with 5 being the best. If you had to play a video game in this world, how satisfied would you be?
    \item What did you like about the experience building this world? What would you change if you could? (freeform response)
    \item Agree or Disagree: The world I have built is interesting
    \item Agree or Disagree: I like the characters I put in the locations
    \item Agree or Disagree: I like the objects I put in the locations
    \item Agree or Disagree: I like how I linked the locations
    \item Agree or Disagree: I like the diversity of locations
    \item Agree or Disagree: I like the diversity of characters
    \item Agree or Disagree: I like the diversity of objects
\end{itemize}

If evaluators had access to model based suggestions, they received these additional questions:
\begin{itemize}
    \item On a scale of 1 to 5, did you find the suggestions in the dropdown menu helpful?
    \item Agree or Disagree: the suggestions made it faster for me to fill in a world
    \item Agree or Disagree: the suggestions were diverse and interesting
    \item Agree or Disagree: I often picked something from the suggestion
    \item Agree or Disagree: I would like having the suggestions again
    \item Did you like the suggestions? Why or why not? (freeform response)
    \item If you could improve the suggestions, how would you do that? (freeform response)
\end{itemize}

\begin{algorithm}[t]
\SetKw{KwConst}{Constant:}
\DontPrintSemicolon
  \caption{Creating a Playable World for LIGHT
  \label{alg:world_creation}
  }
    \KwConst{N = maximum number of locations}\;
    \KwConst{P = filler probability = $0.15$}\;
    \KwConst{X = block percentage}\;
    Initialize an empty grid and fill the center with a randomly selected locations\;
    Set N = maximum number of locations\;
    Block X\% of grid positions\;
    \ForEach{location}{
        \ForEach{location neighbor description}{
            Randomly choose direction for new location\;
            newLocation = PredictLocation(description)\;
            \If{$\mathrm{Random}(0, 1) < P$}{
                newLocation = filler location
            }
            Place newLocation in direction\;
            Prevent newLocation from being predicted again if not filler\;
            Randomly connect newLocation to existing surrounding locations\;
            newCharacters = PredictCharacter(description)\;
            newObjects = PredictObject(description)\;
            \If{num\_locations $\geq N$}{\Return}
        }
    }
\end{algorithm}

\subsection{Generated Maps}

\begin{figure*}[t]
    \centering
    \includegraphics[width=0.7\linewidth]{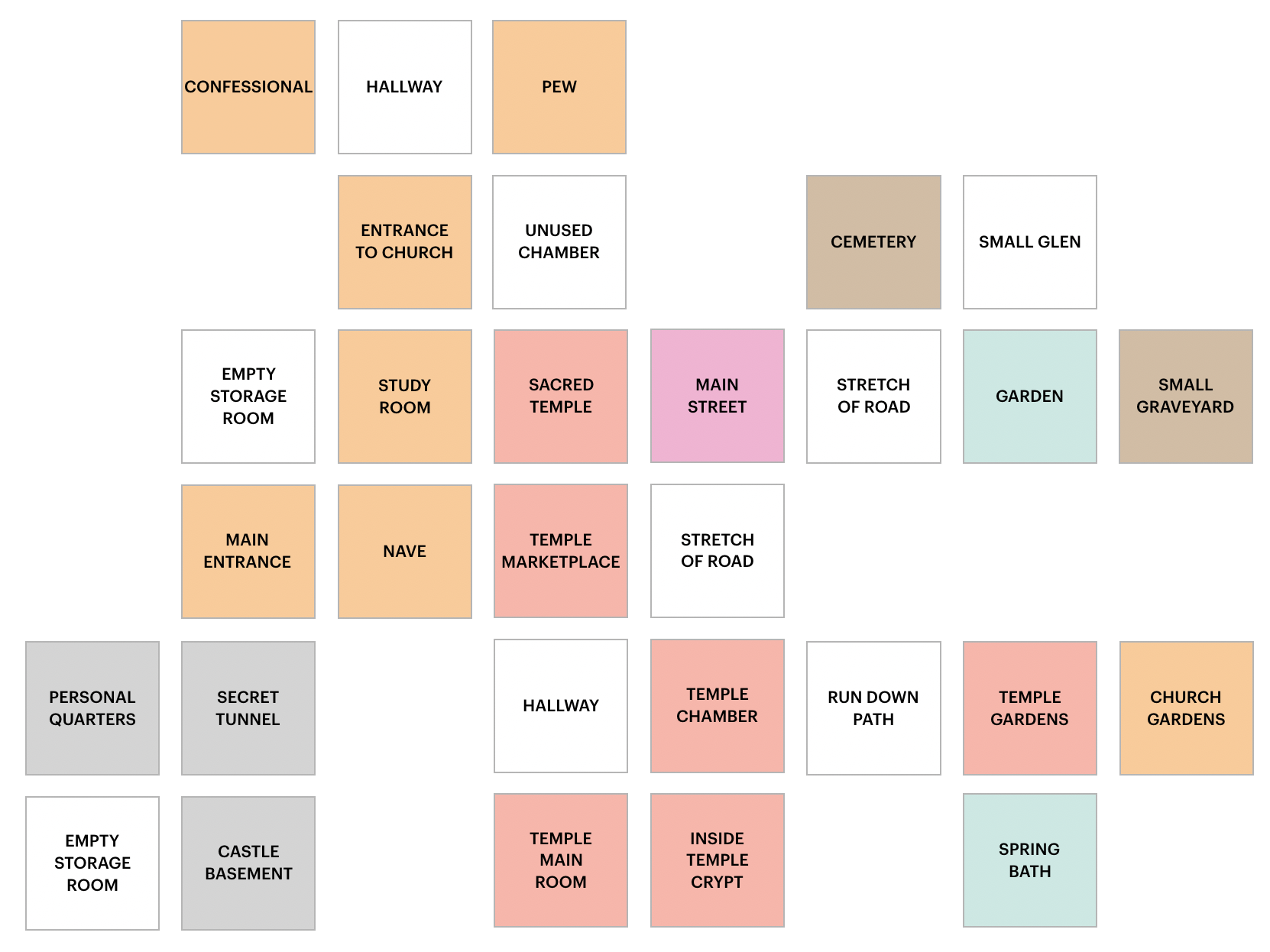}
    \caption{\textbf{Example Generated Map of Linked Locations}}
    \label{fig:example_map1}
\end{figure*}

\begin{figure*}[t]
    \centering
    \includegraphics[width=0.7\linewidth]{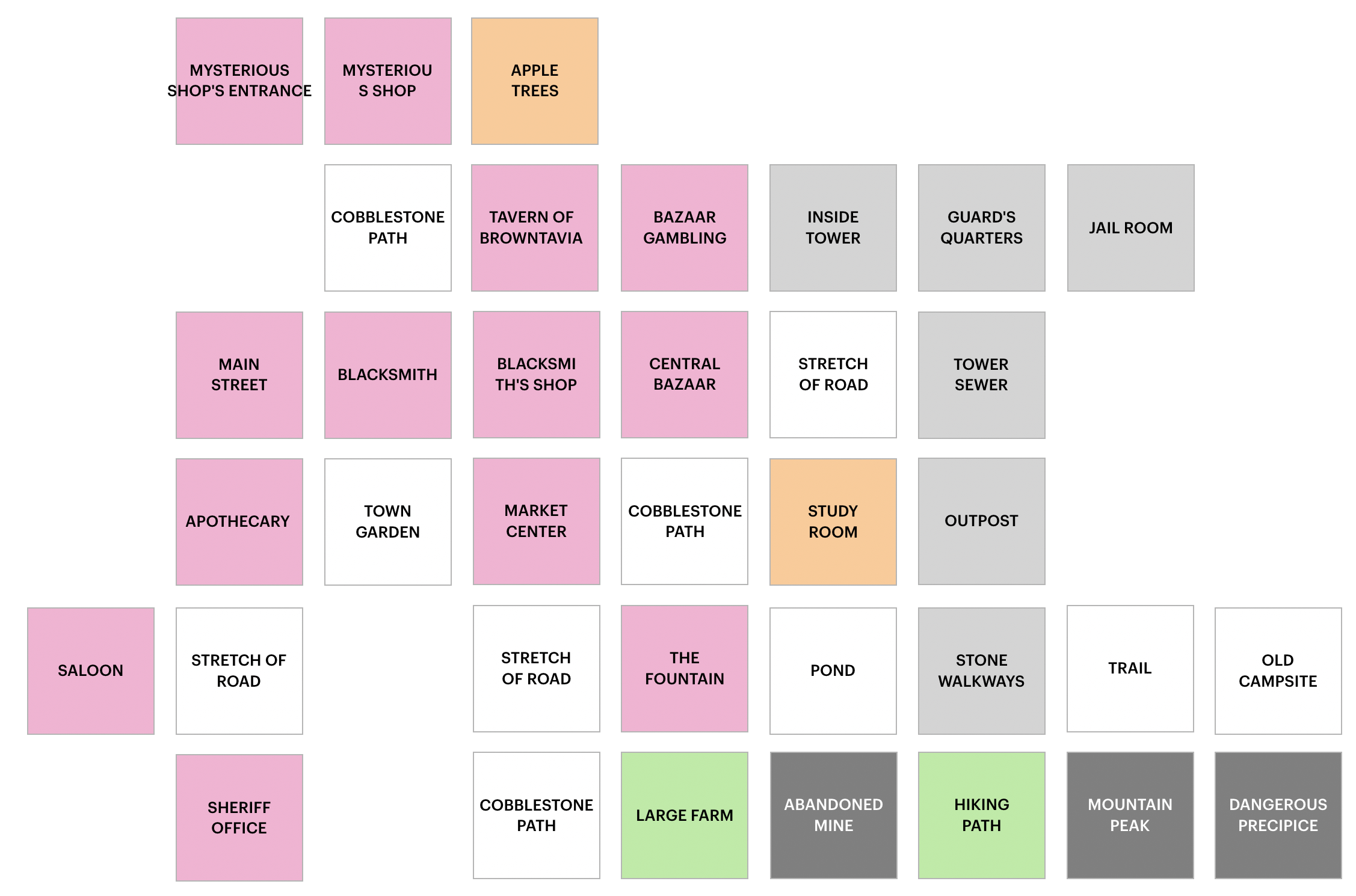}
    \caption{\textbf{Example Generated Map of Linked Locations}}
    \label{fig:example_map2}
\end{figure*}

We show examples of locations arranged by our models in Figure~\ref{fig:example_map1} and Figure~\ref{fig:example_map2}. Filler locations, such as \textit{unused chamber}, \textit{hallway}, and \textit{empty storage room} are shown in white. 

\subsection{Generated Game Elements}

We employ grid search to find the hyperparameters that lead to the best performance. Selected hyperparameters are listed in Table~\ref{tbl:hyperparameters}. We show various additional examples of generated locations, characters, and objects from the best performing models in  Table~\ref{tbl:generated_location_appendix}, Table~\ref{tbl:generated_character_appendix}, and Table~\ref{tbl:generated_object_appendix}.

\begin{table}
  \centering \small 
  \begin{tabular}{l|l|l}
    \toprule
    \textbf{Hyperparameter} & \textbf{Descriptions} & \textbf{Affordances} \\
    \midrule
    \textbf{Number of Layers} & 6 & 6 \\
    \midrule
    \textbf{Embedding Size} & 512 & 512 \\
    \midrule
    \textbf{Dropout} & 0.1 & 0.1 \\
    \midrule
    \textbf{Learning Rate} & $1\mathrm{e}{-4}$ & $7\mathrm{e}{-4}$ \\
    \bottomrule
  \end{tabular}
  \caption{\textbf{Hyperparameters} from models used for generation of game elements}
  \label{tbl:hyperparameters}
\end{table}    

\begin{table}
  \centering \small 
  \begin{tabular}{l|l}
    \toprule
    \textbf{Given Character:} & scientist \\
    \midrule
    \textbf{Generated} & I like to study about magic . the science \\ 
    \textbf{Persona:} & that makes magic works. I work here on \\
    & my own. I work in the lab to make sure \\
    & that other people are paying attention. \\
    \midrule
    \textbf{Generated} & He is a scientist that works in the lab on \\
    \textbf{Description:} & a small scale. He studies magic. \\ 
    \toprule
    \textbf{Given Character:} & general store owner  \\
    \midrule
    \textbf{Generated} & I sell all my goods to the local king. I \\ 
    \textbf{Persona:} & am paid to serve him and i make sure \\
    & he doesn't come in and steal from me. I \\
    & am also a leader amongst the peasants. \\
    \midrule
    \textbf{Generated} & This general store owner is always \\
    \textbf{Description:} & looking for more personal items to buy. \\
    \toprule
    \textbf{Given Character:} & ambassadors \\
    \midrule
    \textbf{Generated} & I am an ambassador of the region. I am \\
    \textbf{Persona:} & not quite sure where I'd like to be \\
    & located , but I will tell them what they \\
    & need to know about my country. I would \\ 
    & like to be able to speak with them , but \\
    & I don't know which one. \\
    \midrule
    \textbf{Generated} & She is a fierce diplomat and seems to \\
    \textbf{Description:} & care deeply about all the people she \\ 
    & meets that are important to them. She is \\
    & a nice girl and has a very kind \\
    & personality. \\
    \toprule
    \textbf{Given Character:} & archaeologists \\
    \midrule
    \textbf{Generated} & I am a archaeologist. I travel to the \\ 
    \textbf{Persona:} & kingdom to gather information. I am a \\
    & historian, and I study what's been \\
    & happening around this kingdom. \\
    \midrule
    \textbf{Generated} & The archaeologist is a hard working \\
    \textbf{Description:} & man and he spends most of its day \\
    & looking for more information to share. \\
    \bottomrule
  \end{tabular}
  \caption{\textbf{Example Generated characters}, given the desired character name.}
  \label{tbl:generated_character_appendix}
\end{table}    

\begin{table}
  \centering \small 
  \begin{tabular}{l|l}
    \toprule
    \textbf{Given Object:} & scientific and writing instruments \\
    \midrule
    \textbf{Generated} & The pen is made of metal and it seems the \\ 
    \textbf{Description:} & hand-written instrument has a unique \\
    & feel in the hand. \\
    \midrule
    \textbf{Generated} & gettable \\
    \textbf{Affordances:} &  \\ 
    \toprule
    \textbf{Given Object:} & ghosts \\
    \midrule
    \textbf{Generated} & The ghost appears to still have its memories \\ 
    \textbf{Description:} & but is a strange looking creature that  \\
    & lives in another world without one. \\
    \midrule
    \textbf{Generated} & None \\
    \textbf{Affordances:} &  \\ 
    \toprule
    \textbf{Given Object:} & mask \\
    \midrule
    \textbf{Generated} & The mask looks old and wrinkled by use. \\ 
    \textbf{Description:} &  \\
    \midrule
    \textbf{Generated} & gettable \\
    \textbf{Affordances:} & wearable \\ 
    \toprule
    \textbf{Given Object:} & weapons \\
    \midrule
    \textbf{Generated} & The weapon is very sturdy, and the iron in \\ 
    \textbf{Description:} & the hilt is very light \\
    \midrule
    \textbf{Generated} & gettable \\
    \textbf{Affordances:} & weapon \\ 
    \bottomrule
  \end{tabular}
  \caption{\textbf{Example Generated Objects}, given the desired object name.}
  \label{tbl:generated_object_appendix}
\end{table}    

\begin{table*}
  \centering \small 
  \begin{tabular}{l|l}
    \toprule
    \textbf{Given Location:} & castle exit \\
    \midrule
    \textbf{Generated Description:} & The entrance to the castle is old and narrow - - it is dark and unbearably dangerous. The exit is \\ 
    & large, with stone walls, and several wooden doors. \\
    \midrule
    \textbf{Generated Background:} & This castle has been abandoned many centuries ago. It is impossible to even escape. The castle is not \\
    & safe for ordinary creatures. \\ 
    \toprule
    \textbf{Given Location:} & Underground Chamber \\
    \midrule
    \textbf{Generated Description:} & The underground chamber is a dark and dark place filled with many deadly creatures. People call it a \\
    & 'cave' but it has been abandoned by the village's inhabitants. No human has crawled out of it. \\
    \midrule
    \textbf{Generated Background:} & This chamber contains information that people of all ages would not have gotten without their help. \\
    & The villagers have a sense of urgency and fear. When they come into the chamber this chamber contains \\
    & information that keeps a secret from humans.\\
    \toprule
    \textbf{Given Location:} & The Pit of Despair \\
    \midrule
    \textbf{Generated Description:} & The pit of despair is a dark place to look at and despair itself is a place where the demons hang out. \\
    & There's even more evil in this corner than there was before. \\
    \midrule
    \textbf{Generated Background:} & The pit of despair is the place where all the evil humans are kept and treated as if they are dead. \\
    & They were found and tortured and the demons are all gone forever. \\
    \toprule
    \textbf{Given Location:} & School \\
    \midrule
    \textbf{Generated Description:} & The school is old and empty; only the last 10 generations had any memories of the past. The local \\
    & schools look the same, but there are a lot of young - gravy looking people in there, showing their talents \\
    & and passion. \\
    \midrule
    \textbf{Generated Background:} & The school's reputation is mostly because it has been established as a place of peace throughout history. \\
    & There are numerous students at each high school that live here in hopes of getting a better education. The \\
    & teachers have also heard of how this school will lead to a higher life . \\
    \bottomrule
  \end{tabular}
  \caption{\textbf{Example Generated Locations}, given the desired object name.}
  \label{tbl:generated_location_appendix}
\end{table*}

\end{document}